\documentclass[11pt]{article}

\usepackage[preprint]{acl}

\usepackage{amsmath,amsfonts,bm}

\def\eqref#1{equation~\ref{#1}}

\def\1{\bm{1}}

\DeclareMathAlphabet{\mathsfit}{\encodingdefault}{\sfdefault}{m}{sl}
\SetMathAlphabet{\mathsfit}{bold}{\encodingdefault}{\sfdefault}{bx}{n}

\DeclareMathOperator*{\argmax}{arg\,max}

\usepackage{times}
\usepackage{enumitem}
\usepackage{latexsym}
\usepackage[T1]{fontenc}
\usepackage[utf8]{inputenc}
\usepackage{microtype}
\usepackage{inconsolata}
\usepackage{enumerate}
\usepackage{graphicx}
\usepackage{booktabs}
\usepackage{multirow}
\usepackage{colortbl}
\usepackage{bbm}
\usepackage{algorithm}
\usepackage{algorithmic}
\usepackage{wrapfig}
\usepackage[most]{tcolorbox}

\usepackage{xcolor}
\definecolor{diffcolor}{RGB}{31, 119, 180}     
\definecolor{rubriccolor}{RGB}{44, 160, 44}    
\definecolor{evalcolor}{RGB}{255, 127, 14}     
\definecolor{judgecolor}{RGB}{148, 103, 189}   

\definecolor{promptbg}{RGB}{245, 245, 250}
\definecolor{promptborder}{RGB}{100, 100, 180}
\newtcolorbox{promptbox}[1][]{%
    enhanced,
    colback=promptbg,
    colframe=promptborder,
    fonttitle=\bfseries,
    title=#1,
    left=5pt,
    right=5pt,
    top=5pt,
    bottom=5pt,
}

\graphicspath{{figures/}}

\title{\methodName{}: Learning Generative Reward Models\\to Judge by the Right Differences}

\newcommand{\methodName}{\textsc{RightDiff}}

\author{
    \textbf{Qiyuan Chen}$^{1}$\thanks{\, Equal Contribution.},
    \textbf{Hongsen Huang}$^{2}$\footnotemark[1],
    \textbf{Jiahe Chen}$^{1}$,\\
    \textbf{Renjie Hua}$^{2,3}$,
    \textbf{Hongxia Xu}$^{1}$\footnotemark[2],
    \textbf{Chuan Ren}$^{2}$\thanks{\, Corresponding Author.},
    \textbf{Jian Wu}$^{1}$\\
    $^{1}$Zhejiang University
    $^{2}$Soochow Securities Co.,Ltd.
    $^{3}$Nanjing University\\
    \texttt{chenqiyuan1012@gmail.com}
}

\begin{document}
\maketitle

\begin{abstract}
Generative reward models typically rely on generic query-level rubrics or unconstrained rationales, overlooking the fact that pairwise preference is driven by response-specific differences. Moreover, optimizing solely for outcome correctness encourages models to exploit superficial shortcuts such as verbosity and formatting, triggering rationale drift. To address this, we propose \methodName{}, which structures pairwise evaluation into an evidence-grounded evaluation trajectory executed in a single autoregressive generation. \methodName{} explicitly discovers contrastive differences between candidates, categorizes them into material or immaterial factors, and induces pair-specific rubrics before final adjudication. To train this process without post-hoc rationalization, we synthesize demonstration data via response-isolated prompts and optimize the reward model with a difference-alignment reward regularized by budget-constrained one-to-one matching and multiplicative gating. \methodName{} achieves state-of-the-art preference modeling performance on \textsc{RewardBench} and \textsc{RM-Bench}, surpassing all open-source reward models and leading proprietary LLM judges. Downstream, it provides reliable supervision for direct preference optimization on \textsc{Arena-Hard} and \textsc{AlpacaEval} while effectively resisting length shortcuts.
\end{abstract}
\section{Introduction}
\label{sec:introduction}

Generative reward models (GenRMs) explicitly generate evaluation rationales or scoring criteria before predicting preferences \citep{zhang2025generative,mahan2024generative,ye2024beyond}, providing a more flexible and interpretable mechanism for preference assessment with large language models \citep{zheng2023judging,yuan2024self}. In particular, rubric-based GenRMs construct evaluation criteria conditioned on the query and then compare candidate responses accordingly \citep{kim2024prometheus,liu2026openrubrics}. However, pairwise preference judgment is inherently relative: what distinguishes two particular responses is often more informative than a generic set of criteria elicited from the query alone, and the final preference may hinge on only a few response-specific differences.

\begin{figure}[t]
    \centering
    \includegraphics[width=\linewidth]{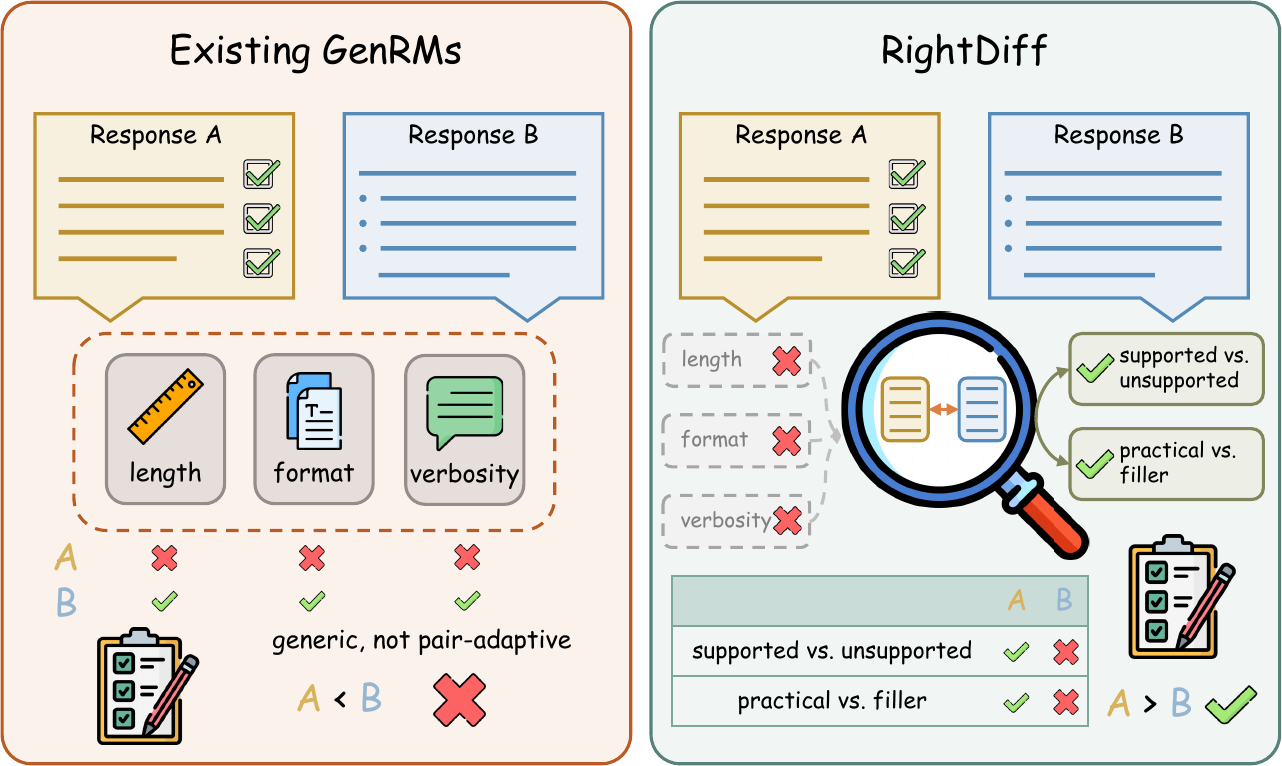}
    \caption{
    Finding the right differences for pairwise reward modeling.
    Existing GenRMs may rely on shallow differences, while \methodName{} identifies preference-relevant differences to construct pair-specific rubrics and reach grounded judgments.
    }
    \label{fig:teaser}
    \vspace{-4mm}
\end{figure}

This raises a dual problem. First, existing methods typically generate rubrics directly from the query or exploit response pairs to construct more discriminative criteria during training \citep{liu2026cdrrm,jia2026open,xu2026alternating,liu2026skywork}. Since the evaluation criteria that matter for a particular comparison can often be determined only after examining the concrete candidates, we argue that difference discovery should be an explicit part of GenRM reasoning at inference time \citep{liu2025inference,huang2026think,chen2026rm}. Second, finding differences is not enough: the model must find the \emph{right} differences. Two responses may differ in many ways, ranging from superficial properties such as length and formatting to substantive differences in factuality, instruction following, or reasoning quality. Under reinforcement learning \citep{ouyang2022training,bai2022constitutional,guo2025deepseek}, when optimization is dominated by final preference correctness, salient but weakly grounded differences may become convenient shortcuts for predicting the label \citep{wang2026outcome}. Worse, in a rubric-generating GenRM, such shallow differences can be further abstracted into seemingly plausible evaluation criteria, yielding a correct preference on the wrong evaluation basis.

This motivates a central question: \emph{among the many differences between two responses, which ones should actually determine the preference?} We refer to these as the \emph{right differences}, formally the preference-relevant differences that provide valid evidence for human judgment. Based on this view, we propose \methodName{} (Figure~\ref{fig:teaser}), which structures pairwise evaluation into a single autoregressive evaluation trajectory consisting of four stages: \emph{difference discovery}, \emph{rubric induction}, \emph{criterion-level evaluation}, and \emph{final adjudication}. Given a query and two candidates, the model identifies their key differences, abstracts them into a pair-specific rubric, evaluates both responses independently under each criterion, and adjudicates the preference from the resulting verdicts. To discourage outcome optimization from reinforcing superficial difference shortcuts, we further introduce \emph{difference-alignment supervision}: an independent evaluator assesses whether the discovered differences and the derived rubric are consistent with human-verified reference evidence, providing an additional training signal beyond final preference correctness \citep{lai2026r,wang2026outcome}.


We systematically evaluate \methodName{} across standard preference benchmarks, including \textsc{RewardBench} and \textsc{RM-Bench}, alongside downstream policy alignment on \textsc{Arena-Hard} and \textsc{AlpacaEval}. Experimental results show that \methodName{} achieves state-of-the-art overall performance, surpassing all open-source reward models, including strong scalar and generative baselines, as well as leading proprietary LLM judges. This advantage is especially pronounced on subtle and distractor-heavy comparisons such as \textsc{RewardBench} \texttt{Chat-Hard} and \textsc{RM-Bench} \texttt{Chat}. \methodName{} also provides reliable supervision for downstream direct preference optimization (DPO) \citep{rafailov2023direct}, significantly improving open-ended instruction following as measured by win rates while remaining robust against superficial length shortcuts. Further process-level evaluations and training dynamics confirm that our method mitigates rationale drift by anchoring decisions in verifiable evidence.


In summary, our contributions are threefold:
\begin{itemize}[leftmargin=*,noitemsep,topsep=2pt]
  \item We formalize pairwise reward modeling around discovering preference-relevant differences, explicitly distinguishing material differences from superficial stylistic variations.
  
  \item We propose \methodName{}, a structured framework that unifies difference discovery, rubric induction, criterion-level evaluation, and final adjudication within a single autoregressive generation during inference.
  
  \item We design a difference-alignment reward, regularized by budget-constrained one-to-one matching and multiplicative gating, that suppresses superficial shortcut exploitation and improves downstream policy alignment.
\end{itemize}

\begin{figure*}[t]
\centering
\includegraphics[width=0.95\textwidth]{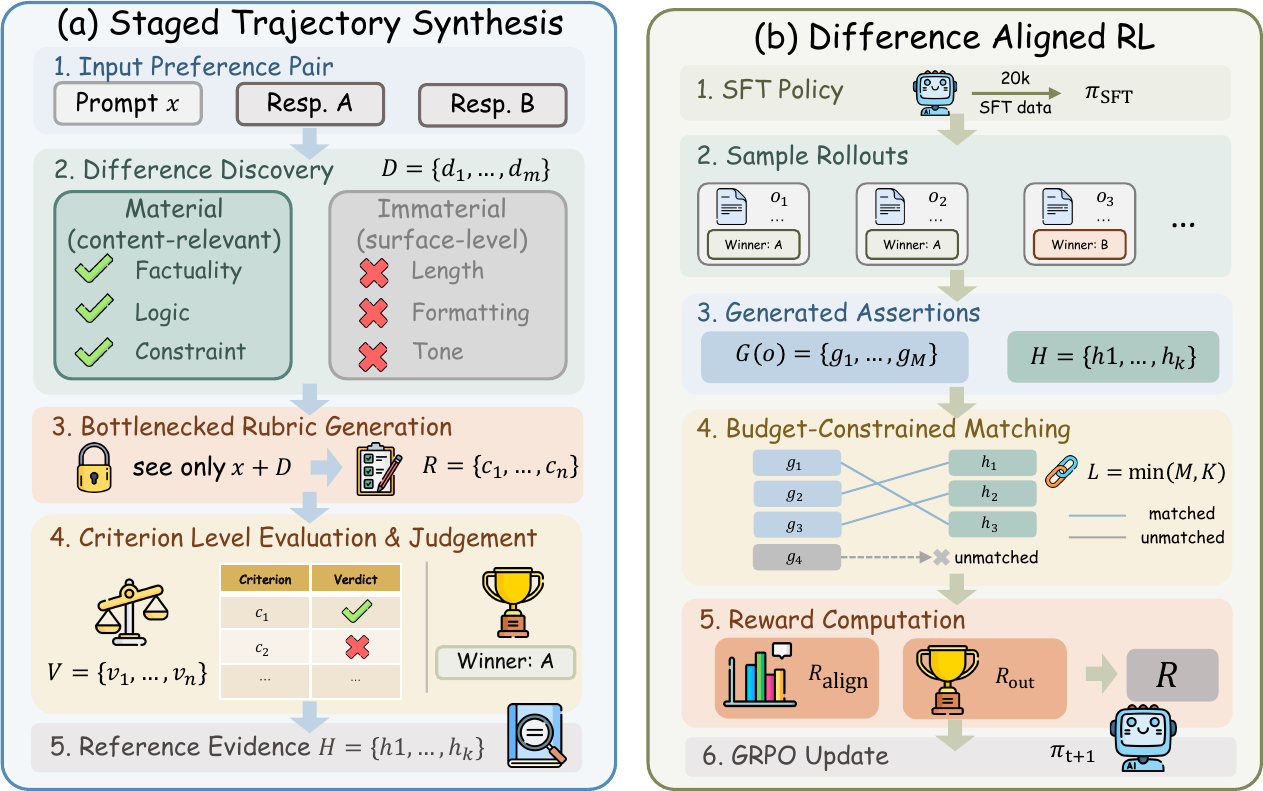}
\caption{
Overview of \methodName{}. 
\textbf{(a) Staged Data Synthesis}: We construct cold-start demonstration trajectories through four sequential, decoupled steps. By isolating candidate responses from the rubric generator, the model is forced to derive evaluation criteria $\mathcal{R}$ strictly from discovered material differences $\mathcal{D}$. Verified trajectories provide reference evidence $\mathcal{H} = (h_1, \ldots, h_K)$ for subsequent training. 
\textbf{(b) Difference-Aligned Reinforcement Learning}: The reward model is trained with Group Relative Policy Optimization (GRPO) using an outcome-aligned dual reward. A lightweight MetaJudge pairs generated claims with reference evidence using one-to-one matching within a capacity budget $L = \min(M, K)$, penalizing verbosity and vague claims. The alignment reward $R_{\mathrm{align}}$ is multiplicatively gated with outcome correctness $R_{\mathrm{out}}$ so that positive advantage is granted only when the model reaches the correct judgment with evidence aligned to the reference.
}
\label{fig:method}
\vspace{-4mm}
\end{figure*}

\section{Method}
\label{sec:method}

\methodName{} turns pairwise reward modeling into a structured, four-stage reasoning process. Given a prompt $x$ and candidate responses $(y^A, y^B)$, the model generates an evaluation trajectory demarcated by explicit semantic tags in a single autoregressive generation:
\begin{equation}
\label{eq:trajectory}
\small
\begin{split}
\tau = \Big( 
& \textcolor{diffcolor}{\texttt{<|diff|>}} \, \mathcal{D} \, \textcolor{diffcolor}{\texttt{<|/diff|>}} \rightarrow \\
& \textcolor{rubriccolor}{\texttt{<|rubric|>}} \, \mathcal{R} \, \textcolor{rubriccolor}{\texttt{<|/rubric|>}} \rightarrow \\
& \textcolor{evalcolor}{\texttt{<|eval|>}} \, \mathcal{E} \, \textcolor{evalcolor}{\texttt{<|/eval|>}} \rightarrow \\
& \textcolor{judgecolor}{\texttt{<|judge|>}} \, \mathcal{J} \, \hat{y} \, \textcolor{judgecolor}{\texttt{<|/judge|>}} \Big).
\end{split}
\end{equation}
The evaluation proceeds across four sequential stages. 
In \textcolor{diffcolor}{\textbf{\texttt{<|diff|>}}}, the model identifies concrete differences $\mathcal{D} = \{d_1, \ldots, d_n\}$ between $y^A$ and $y^B$, explicitly separating material differences from immaterial surface features. 
In \textcolor{rubriccolor}{\textbf{\texttt{<|rubric|>}}}, the model converts these material differences into an objective rubric $\mathcal{R} = \{c_1, \ldots, c_m\}$. 
In \textcolor{evalcolor}{\textbf{\texttt{<|eval|>}}}, the model evaluates both responses independently under each criterion in $\mathcal{R}$ to produce local verdicts $\mathcal{E} = \{v_1, \ldots, v_m\}$. 
Finally, in \textcolor{judgecolor}{\textbf{\texttt{<|judge|>}}}, the model weighs these criterion-level evaluations to produce a concise justification $\mathcal{J}$ and select the winning response $\hat{y} \in \{A, B\}$.

As illustrated in Figure~\ref{fig:method}, our framework consists of two main training stages. 
First, we synthesize cold-start demonstration trajectories using an isolated teacher pipeline that deliberately hides candidate responses when inducing rubrics, so that criteria are derived from the discovered differences rather than post-hoc justifications. 
Second, we optimize the model with reinforcement learning using a difference-alignment reward that verifies whether generated differences match the reference evidence, discouraging superficial shortcuts. 
Although data synthesis decouples these steps to improve demonstration quality (Figure~\ref{fig:method}(a)), our trained model executes the entire trajectory in Eq.~(\ref{eq:trajectory}) within a single autoregressive generation during inference.

\subsection{Synthesizing Training Data via Response-Isolated Prompts}
\label{sec:data_synthesis}

When an LLM generates evaluation criteria while viewing both candidate responses, it often constructs criteria post-hoc simply to defend an intuitive preference. 
To mitigate this bias, we synthesize training demonstrations by separating the teacher generation into four sequential steps, as depicted in Figure~\ref{fig:method}(a). 
The ground-truth preference label $y^\star$ is withheld throughout generation and used solely for post-hoc quality filtering.

\paragraph{Step 1: Difference Discovery.}
Given instruction $x$ and candidates $(y^A, y^B)$, the teacher extracts a concise set of pair-specific differences $\mathcal{D} = \{d_1, \ldots, d_n\}$. 
Each difference $d_i = \langle \kappa_i, e_i^A, e_i^B, \ell_i \rangle$ specifies the comparative aspect $\kappa_i$, quotes supporting textual spans $e_i^A$ and $e_i^B$, and assigns a materiality label $\ell_i \in \{\texttt{Material}, \texttt{Immaterial}\}$. 
A difference is labeled as \texttt{Material} if it substantively influences factual correctness, reasoning validity, or explicit constraint fulfillment. 
Stylistic variations such as response length, decorative formatting, or superficial filler are labeled as \texttt{Immaterial} unless specifically requested by the prompt. 
The teacher is instructed not to issue preference judgments at this step, teaching the model to acknowledge surface variations without treating them as preference evidence.

\paragraph{Step 2: Rubric Induction with Response Isolation.}
If the teacher can inspect candidate responses $(y^A, y^B)$ while drafting criteria, it can bypass the discovered differences $\mathcal{D}$ and write rules directly around salient surface features. 
To block direct access to the full response texts during rubric induction, we exclude $(y^A, y^B)$ from the rubric prompt:
\begin{equation}
\mathcal{R} = \mathcal{T}_{\mathrm{rubric}}(x, \mathcal{D}),
\end{equation}
where $\mathcal{T}_{\mathrm{rubric}}$ denotes the teacher's rubric-induction module. 
Because the teacher only observes prompt $x$ and differences $\mathcal{D}$, it is forced to derive criteria $\mathcal{R} = \{c_1, \ldots, c_m\}$ strictly by abstracting the identified material differences. 
Criteria must satisfy two formal properties: evidence traceability requires every criterion to map directly to a material difference in $\mathcal{D}$, and answer agnosticism requires criteria to describe objective requirements without referencing candidate identities or revealing the preferred response.

\paragraph{Step 3: Criterion-Level Evaluation.}
With rubric $\mathcal{R}$ fixed, candidate responses are reintroduced into the evaluation prompt. 
The model assesses both candidates against each criterion independently, producing local verdicts $\{v_1, \ldots, v_m\}$ where each verdict $v_j \in \{A, B, \texttt{Tie}\}$ is grounded in explicit reasoning. 
No criteria can be added, deleted, or modified during this step.

\paragraph{Step 4: Final Adjudication and Quality Filtering.}
Finally, the teacher aggregates local verdicts into an overall evaluation:
\begin{equation}
(\mathcal{J}, \hat{y}) = \mathcal{T}_{\mathrm{judge}}(x, y^A, y^B, \mathcal{D}, \mathcal{R}, \mathcal{E}),
\end{equation}
where $\mathcal{T}_{\mathrm{judge}}$ denotes the teacher's adjudication module. 
Instead of relying on unweighted majority voting, the teacher resolves conflicts by weighting violation severities, producing an integrated rationale $\mathcal{J}$ and final prediction $\hat{y}$. 
We discard any trajectory where the predicted winner disagrees with the ground-truth human label or where quoted snippets cannot be verified verbatim in the responses. 
For each verified trajectory, its material differences and corresponding criteria are serialized into an ordered reference evidence list $\mathcal{H} = (h_1, \ldots, h_K)$, providing supervision for reinforcement learning. The complete prompts for the four synthesis steps are provided in Appendix~\ref{app:prompts} (Figures~\ref{fig:prompt_teacher_contrast}--\ref{fig:prompt_teacher_judgment}).

\subsection{Reinforcement Learning with Difference Alignment}
\label{sec:rl_alignment}

Supervised fine-tuning (SFT) initializes the model with the syntax of structured evaluation. 
However, optimizing solely for final preference correctness triggers rationale drift: the model quickly learns that superficial cues such as response length or formatting can predict the label, prompting it to fabricate plausible-sounding rubrics to justify those shortcuts. 
To anchor exploration to verified evidence, we optimize \methodName{} via GRPO \citep{shao2024deepseekmath} regularized by a difference-alignment reward that verifies generated reasoning against reference evidence, as shown in Figure~\ref{fig:method}(b).

\paragraph{One-to-One Matching with a Capacity Budget.}
During reinforcement learning rollouts, the model produces, for each sampled completion $o$, a sequence of generated difference and rubric claims $\mathcal{G}(o) = (g_1, \ldots, g_M)$, which are compared against reference evidence $\mathcal{H} = (h_1, \ldots, h_K)$. 
Here each claim $g_i$ corresponds to either a declared difference or a rubric criterion, so $M$ denotes the total number of claims. 
Unconstrained semantic matching invites two common reward hacking behaviors: the model can output a single vague statement to match multiple reference items, or generate dozens of candidate differences to maximize match coverage through brute-force enumeration.

To block both exploits, we introduce two structural constraints illustrated in Figure~\ref{fig:method}(b). 
First, we apply a capacity budget $L = \min(M, K)$ and retain only the first $L$ assertions:
\begin{equation}
\label{eq:budget}
\widetilde{\mathcal{G}}(o) = (g_1, \ldots, g_L).
\end{equation}
This truncation removes any incentive to pad completions with extraneous differences.
Second, budgeted claims and reference evidence are evaluated by a lightweight MetaJudge in a single autoregressive generation. 
The MetaJudge enforces strict one-to-one matching, mapping each generated claim $g_i$ to at most one reference item $h_j$ with semantic match score $s_{ij} \in [0, 1]$. 
This matching problem is formalized as:
\begin{equation}
\label{eq:matching}
\sigma^\star = \argmax_{\sigma \in \Sigma} \sum_{(i, j) \in \sigma} s_{ij},
\end{equation}
where $\Sigma$ denotes the set of all injective matchings between budgeted assertions and reference items.

\paragraph{Gated Reward Formulation.}
The difference-alignment reward $R_{\mathrm{align}}(o)$ is computed as the normalized matching score over reference count $K$:
\begin{equation}
\label{eq:align_reward}
R_{\mathrm{align}}(o) = \frac{1}{K} \sum_{(i, j) \in \sigma^\star} s_{ij}.
\end{equation}
Normalizing by $K$ penalizes omitted evidence, directly rewarding high coverage of substantive preference determinants.

To integrate preference correctness with process alignment, we gate the rewards multiplicatively:
\begin{equation}
\label{eq:hybrid_reward}
R(o) = R_{\mathrm{out}}(o) \cdot R_{\mathrm{align}}(o),
\end{equation}
where $R_{\mathrm{out}}(o) = \mathbbm{1}[\hat{y} = y^\star]$ indicates whether the final winner is correct. 
Incorrect verdicts thus receive zero reward, and positive advantage accrues only to rollouts that reach the correct judgment through verified, preference-relevant differences.

\paragraph{Policy Optimization via GRPO.}
We optimize \methodName{} with GRPO \citep{shao2024deepseekmath}. 
For each prompt $x$, the model samples $G$ rollouts $\{o_1, \ldots, o_G\} \sim \pi_{\theta_{\mathrm{old}}}$ and obtains scalar rewards $r_i = R(o_i)$. 
Relative advantages within the group are normalized as:
\begin{equation}
\label{eq:grpo_advantage}
\widehat{A}_i = \frac{r_i - \operatorname{mean}(\{r_1, \ldots, r_G\})}{\operatorname{std}(\{r_1, \ldots, r_G\}) + \epsilon},
\end{equation}
where $\epsilon > 0$ is a small constant added for numerical stability. 
The policy parameters $\theta$ are updated by maximizing:
\begin{multline}
\label{eq:grpo_objective}
\mathcal{O}(\theta) = \mathbb{E}\Bigg[ \frac{1}{G} \sum_{i=1}^G \Big( \frac{\pi_\theta(o_i \mid x)}{\pi_{\theta_{\mathrm{old}}}(o_i \mid x)} \widehat{A}_i \\
- \beta D_{\mathrm{KL}}\big(\pi_\theta(\cdot \mid x) \,\|\, \pi_{\mathrm{ref}}(\cdot \mid x)\big) \Big) \Bigg],
\end{multline}
where the expectation is taken over prompts and grouped rollouts, $\beta$ controls the strength of the KL regularization, and $\pi_{\mathrm{ref}}$ is a frozen reference policy, which we instantiate as the SFT checkpoint. 
Under pure outcome supervision, all completions reaching the correct label receive identical rewards, resulting in vanishing intra-group variance ($\widehat{A}_i \approx 0$). 
Our gated formulation restores informative gradient variance across correct answers, reinforcing reasoning that better matches the reference evidence while suppressing shortcut exploration.
\section{Experiments}
\label{sec:experiments}

We evaluate \methodName{} to answer four core research questions:
\begin{itemize}[leftmargin=*,noitemsep,topsep=1pt]
  \item \textbf{RQ1 (Multi-Domain Preference Modeling):} Does \methodName{} excel in standard multi-domain preference evaluations? (\S\ref{sec:rq1})
  \item \textbf{RQ2 (Downstream Policy Alignment):} Can \methodName{} effectively guide downstream policy optimization and improve open-ended instruction following? (\S\ref{sec:rq2})
  \item \textbf{RQ3 (Reinforcement Learning Formulation and Constraint Ablation):} How do the reasoning pipeline components and the difference-alignment reward design contribute to overall performance? (\S\ref{sec:rq3})
  \item \textbf{RQ4 (Human Consensus and Process Fidelity):} Does \methodName{} capture the consensus differences that humans actually use to form preferences on \textsc{HelpSteer3}? (\S\ref{sec:rq4})
\end{itemize}

\subsection{Experimental Setup}
\label{sec:setup}

\paragraph{Training Data.}
We collect approximately 109K preference pairs from Skywork Reward Preference 80K~\citep{liu2026skywork}, \textsc{HelpSteer3}~\citep{wang2025helpsteer3}, Math-DPO-10K~\citep{lai2024step}, and Code-Preference-Pairs~\citep{lai2026r}, and use \texttt{Qwen3.5-397B-A17B} to annotate each pair with a structured evaluation, keeping only examples whose predicted preference matches the original label. We use 50K examples for SFT. On the remaining data, we sample 8 judgments per pair from the SFT reward model and retain comparisons that are answered correctly in 2--6 out of 8 judgments, i.e., pairs near the model's decision boundary; from this pool we sample 5K comparisons for reinforcement learning. A detailed description of the four data sources is provided in Appendix~\ref{appen_sub:training_data}.

\paragraph{Implementation Details.}
We use \texttt{Qwen3-4B} and \texttt{Qwen3-8B} as backbone models and fully fine-tune both. We first perform SFT for one epoch with a learning rate of $1\times10^{-5}$ and a global batch size of 64, and then optimize with RL for 200 steps using $G=8$ rollouts per prompt, a learning rate of $1\times10^{-6}$, and a maximum completion length of 4,096 tokens. All experiments are conducted on 16 NVIDIA A100 GPUs. The full training configuration is described in Appendix~\ref{appen_sub:impl}.

\begin{table*}[!t]
\caption{
\textbf{Pairwise preference evaluation results on RewardBench and RM-Bench across diverse domains.} 
\textbf{Bold} and \underline{underlined} values denote the best and second-best results in each column, respectively (ties are all marked). 
``Avg.'' represents the macro-average of the overall accuracy across the two benchmarks. 
\methodName{} surpasses all open-source generative and scalar reward models in terms of the overall average, and outperforms proprietary LLM judges on both the overall average and subtlety-focused evaluations.
}
\label{tab:main_results}
\centering
\footnotesize
\resizebox{\textwidth}{!}{%
\begin{tabular}{@{}l ccccc ccccc c@{}}
\toprule
\multirow{2}{*}{\textbf{Models}} & \multicolumn{5}{c}{\textbf{RewardBench}} & \multicolumn{5}{c}{\textbf{RM-Bench}} & \multirow{2}{*}{\textbf{Avg.}} \\
\cmidrule(lr){2-6} \cmidrule(lr){7-11}
& \textbf{Chat} & \textbf{Chat-Hard} & \textbf{Safety} & \textbf{Reason.} & \textbf{Overall} & \textbf{Chat} & \textbf{Math} & \textbf{Code} & \textbf{Safety} & \textbf{Overall} & \\
\midrule
\multicolumn{12}{c}{\textit{\textbf{LLM-as-a-Judge}}} \\
\midrule
\texttt{Claude-3-5-sonnet-20240620} & 96.4 & 74.0 & 81.6 & 84.7 & 84.2 & 62.5 & 62.6 & 54.4 & 64.4 & 61.0 & 72.6 \\
\texttt{Llama-3.1-70B-Instruct} & 97.2 & 70.2 & 82.8 & 86.0 & 84.0 & 64.3 & 67.3 & 47.5 & 83.0 & 65.5 & 74.8 \\
\texttt{GPT-4o-0806} & 96.1 & 76.1 & 86.6 & 88.1 & 86.7 & 67.2 & 67.5 & 63.6 & 91.7 & 72.5 & \underline{79.6} \\
\midrule
\multicolumn{12}{c}{\textit{\textbf{Scalar Reward Models}}} \\
\midrule
\texttt{SteerLM-RM-70B} & 91.3 & 80.3 & \textbf{92.8} & 90.6 & \underline{88.8} & 56.4 & 53.0 & 49.3 & 51.2 & 52.5 & 70.7 \\
\texttt{Eurus-RM-7b} & \underline{98.0} & 65.6 & 81.4 & 86.3 & 82.8 & 59.9 & 60.2 & 56.9 & 86.5 & 65.9 & 74.4 \\
\texttt{InternLM2-7B-Reward} & \textbf{99.2} & 69.5 & 87.2 & \underline{94.5} & 87.6 & 61.7 & 71.4 & 49.7 & 85.5 & 67.1 & 77.4 \\
\texttt{ArmoRM-Llama3-8B-v0.1} & 96.9 & 76.8 & \underline{90.5} & \textbf{97.3} & \textbf{90.4} & 67.8 & 57.5 & 53.1 & \underline{92.4} & 67.7 & 79.1 \\
\midrule
\multicolumn{12}{c}{\textit{\textbf{Generative Reward Models}}} \\
\midrule
\texttt{RubricARM-Qwen3-8B} & 90.8 & 81.0 & 59.3 & 82.2 & 78.3 & 69.8 & 69.2 & 63.4 & 62.3 & 66.2 & 72.2 \\
\texttt{RubricARROW-Qwen3-8B} & 90.2 & \underline{81.1} & 63.8 & 79.1 & 78.6 & 67.7 & 71.0 & \textbf{66.6} & 71.4 & 69.2 & 73.9 \\
\texttt{RubricRM-Qwen3-8B} & 92.2 & 76.2 & 83.9 & 78.4 & 82.7 & 61.6 & 69.9 & 59.2 & 89.2 & 70.0 & 76.3 \\
\texttt{RM-R1-DeepSeek-Distilled-7B} & 88.9 & 66.2 & 78.4 & 87.0 & 80.1 & 64.0 & \underline{83.9} & 56.2 & 85.3 & 72.4 & 76.3 \\
\texttt{RM-R1-Qwen2.5-7B-Instruct} & 94.1 & 74.6 & 85.2 & 86.7 & 85.2 & 66.6 & 67.0 & 54.6 & \textbf{92.6} & 70.2 & 77.7 \\
\texttt{FARE-Qwen3-8B} & 96.5 & 70.7 & 85.7 & 89.2 & 85.5 & 62.0 & 71.6 & 59.2 & 89.4 & 70.5 & 78.0 \\
\texttt{JudgeLRM-Qwen3-8B} & 96.6 & 60.5 & 77.8 & 93.7 & 82.2 & 54.2 & \textbf{86.4} & \underline{65.5} & 91.3 & \underline{74.4} & 78.3 \\
\midrule
\multicolumn{12}{c}{\textit{\textbf{Ours}}} \\
\midrule
\rowcolor{purple!10}
\methodName-Qwen3-4B & 89.7 & 81.0 & 84.7 & 84.2 & 84.8 & \underline{71.2} & 73.8 & 60.3 & 91.0 & 74.1 & 79.5 \\
\rowcolor{purple!10}
\methodName-Qwen3-8B & 89.4 & \textbf{82.6} & 86.0 & 86.2 & 86.1 & \textbf{73.1} & 75.9 & 63.2 & \underline{92.4} & \textbf{76.1} & \textbf{81.1} \\
\bottomrule
\end{tabular}
\vspace{-4mm}
}
\end{table*}

\paragraph{Evaluation Benchmarks and Metrics.}
We evaluate \methodName{} as a pairwise reward model on two widely used reward-modeling benchmarks: \textsc{RewardBench}~\citep{lambert2025rewardbench} and \textsc{RM-Bench}~\citep{liu2025rm}. \textsc{RewardBench} measures preference discrimination across \texttt{Chat}, \texttt{Chat-Hard}, \texttt{Safety}, and \texttt{Reasoning}. \textsc{RM-Bench} probes sensitivity to response content while penalizing reward models that can be swayed by stylistic bias, and reports \texttt{Chat}, \texttt{Math}, \texttt{Code}, and \texttt{Safety} separately. We follow the official splits and scoring rules of each benchmark, report accuracy (\%) per domain, and average the two overall scores. Detailed benchmark descriptions and the evaluation baselines are provided in Appendices~\ref{appen_sub:benchmark} and~\ref{appen_sub:baselines}.

\subsection{RQ1: Multi-Domain Preference Modeling Performance}
\label{sec:rq1}

As summarized in Table~\ref{tab:main_results}, \methodName{} achieves the best macro average across both \textsc{RewardBench} and \textsc{RM-Bench}, surpassing all open-source reward models, including leading scalar and generative baselines, and outperforming proprietary LLM judges on the overall average (81.1 vs. 79.6 for the strongest proprietary judge). This advantage is particularly evident on challenging subsets involving subtle quality disparities and stylistic distractors, such as \textsc{RewardBench} \texttt{Chat-Hard} and \textsc{RM-Bench} \texttt{Chat}, where \methodName{} sets a new state of the art. Prior generative reward models typically rely on generic query-level rubrics or unconstrained critiques that are vulnerable to superficial noise. In contrast, explicitly discovering contrastive differences enables \methodName{} to isolate decisive comparative evidence during inference. In addition, while strong baselines frequently suffer from cross-domain imbalances by sacrificing conversational alignment or safety for specialized mathematical reasoning, \methodName{} maintains robust and consistent accuracy across reasoning, coding, dialogue, and safety tasks. Notably, even our 4B variant surpasses all open-source 7B and 8B baselines on the overall average and rivals the strongest proprietary judge (79.5 vs. 79.6). This confirms that structuring pairwise evaluation into explicit difference-driven reasoning introduces a strong inductive bias, substantially reducing reliance on sheer parameter scale.

\begin{table}[t]
\centering
\caption{
Comparison of different strategies applied to \texttt{Qwen2.5-7B-Instruct} on \textsc{Arena-Hard} and \textsc{AlpacaEval}. 
\textbf{WC} denotes the win rate (against \texttt{GPT-4-0314} on \textsc{Arena-Hard}), and \textbf{LC} denotes the length-controlled win rate on \textsc{AlpacaEval}. 
\textbf{Bold} and \underline{underline} indicate the best and second-best results in each column.
}
\label{tab:alpaca}
\resizebox{\linewidth}{!}{%
\arrayrulecolor{black!50}
\begin{tabular}{l|c|cc|c}
\toprule
\multirow{2}{*}{\textbf{Model}} 
& \multicolumn{1}{c}{\textbf{Arena-Hard}} 
& \multicolumn{2}{c|}{\textbf{AlpacaEval}} 
& \multirow{2}{*}{\textbf{AVG}} \\ 
\cmidrule(lr){2-2} \cmidrule(lr){3-4}
& \textbf{WC} 
& \textbf{WC} & \textbf{LC} &  \\ 
\midrule
\texttt{GPT-4-0314}           & 50.0 & 22.1 & 35.3 & 35.8 \\ 
\midrule
\texttt{Qwen2.5-7B-Instruct} & 51.3 & 33.5 & 36.2 & 40.3 \\
+ DPO (via \texttt{Skywork-RM})  & \underline{55.1} & 44.8 & 41.5 & \underline{47.1} \\
+ DPO (via \texttt{ArmoRM})   & 50.8 & 37.6 & 38.1 & 42.2 \\
+ DPO (via \texttt{UltraFeedback}) & 52.8 & 33.7 & 38.7 & 41.7 \\
+ DPO (via \texttt{AI Judge}) & 51.0 & 28.8 & 33.4 & 37.7 \\
+ DPO (via \texttt{RLCF})     & 54.6 & 36.2 & 37.1 & 42.6 \\ 
+ DPO (via \texttt{RubricRM}) & 52.9 & \underline{47.0} & 41.3 & \underline{47.1} \\
\midrule
\rowcolor{purple!10}
+ DPO (via \methodName-4B)            & 53.8 & 45.8 & \underline{41.6} & \underline{47.1} \\ 
\rowcolor{purple!10}
+ DPO (via \methodName-8B)            & \textbf{56.4} & \textbf{47.1} & \textbf{42.5} & \textbf{48.7} \\
\bottomrule
\end{tabular}%
}
\vspace{-4mm}
\end{table}

\subsection{RQ2: Downstream Policy Alignment}
\label{sec:rq2}

As shown in Table~\ref{tab:alpaca}, the discriminative superiority of \methodName{} directly translates into measurable policy improvements when used to supervise downstream DPO~\citep{rafailov2023direct}. The policy model aligned with \methodName-8B achieves the highest overall win rates on both \textsc{Arena-Hard} and \textsc{AlpacaEval}, outperforming all existing reward model baselines. Moreover, the policy optimized via \methodName{} achieves the top length-controlled (LC) win rate on \textsc{AlpacaEval}, whereas baselines often exploit verbosity as an easy shortcut and drop noticeably once length is strictly controlled, indicating our resistance to such superficial bias. Because \methodName{} explicitly designates superficial length and decorative formatting as immaterial differences during preference adjudication, it provides optimization gradients focused on substantive content, effectively limiting verbosity inflation in the downstream policy. In addition, our 4B reward model delivers downstream alignment results comparable to much larger baseline models, confirming that difference-grounded evaluation provides stable and robust supervision for policy post-training.

\begin{table}[t]
\centering
\caption{
\textbf{Ablation of reinforcement learning formulations and constraints on \texttt{Qwen3-8B}.} 
``S-Corr'' denotes Spurious Correctness, and Avg. \# Diff is the average number of generated differences per pair.
}
\label{tab:rl_ablation}
\resizebox{\linewidth}{!}{%
\begin{tabular}{l|cc|cc}
\toprule
\textbf{Configuration} & \textbf{RewardBench} & \textbf{RM-Bench} & \textbf{S-Corr (\%)} $\downarrow$ & \textbf{Avg. \# Diff} \\
\midrule
$\pi_{\mathrm{SFT}}$ & 84.2 & 73.5 & 11.5 & 3.2 \\
\quad + Outcome-only & 85.4 & 71.8 & 21.6 & 2.5 \\
\quad + Additive & 85.1 & 74.0 & 10.2 & 3.8 \\
\quad + $w/o$ Budget & 85.8 & 75.2 & 6.8 & 8.1 \\
\quad + $w/o$ 1-to-1 & 85.5 & 74.5 & 9.4 & 3.5 \\
\midrule
\rowcolor{purple!10}
\textbf{\methodName{}} & \textbf{86.1} & \textbf{76.1} & \textbf{5.1} & \textbf{3.3} \\
\bottomrule
\end{tabular}%
}
\end{table}

\subsection{RQ3: Reinforcement Learning Formulation and Constraint Ablation}
\label{sec:rq3}

\begin{figure}[t]
\centering
\includegraphics[width=\linewidth]{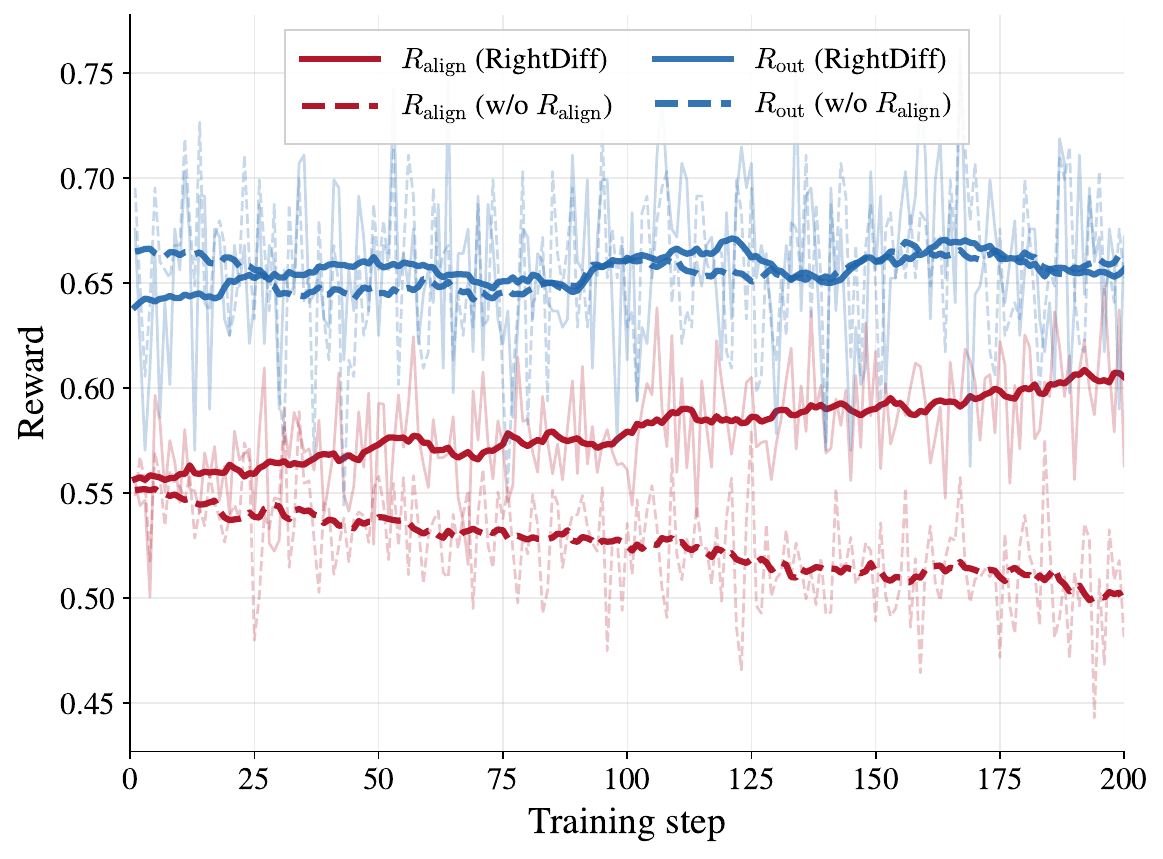}
\caption{
\textbf{Training dynamics of reward optimization under different supervision signals across 200 training steps.} 
Solid lines denote \methodName{} and dashed lines denote the outcome-only variant without the difference-alignment reward ($w/o$ $R_{\mathrm{align}}$). 
Faint curves represent raw batch metrics, while bold curves represent smoothed trends. 
While both configurations achieve comparable outcome rewards ($R_{\mathrm{out}}$), removing the difference-alignment reward causes a persistent decline in $R_{\mathrm{align}}$, validating that pure outcome optimization leads to rationale drift.
}
\label{fig:training_dynamics}
\end{figure}

To see whether difference alignment reduces shortcut behavior and whether our matching rules block reward-hacking paths, we test different RL designs in Table~\ref{tab:rl_ablation} and track their training in Figure~\ref{fig:training_dynamics}.

\paragraph{Training Dynamics and Policy Degeneration.}
Training solely on final win-or-lose labels causes rationale drift. As Figure~\ref{fig:training_dynamics} shows, both models reach similar outcome rewards, but the alignment score of the outcome-only model steadily falls: without rationale supervision, the model guesses from surface cues rather than comparing responses carefully. Table~\ref{tab:rl_ablation} shows the consequence: superficial shortcuts stay competitive on simple benchmarks but fail on harder ones, sharply increasing spurious correctness (S-Corr), i.e., correct verdicts reached without reference-aligned evidence.

\paragraph{Reward Formulation and Structural Constraints.}
In Table~\ref{tab:rl_ablation}, the simple additive reward underperforms because the model still receives reward for plausible commentary even when it picks the wrong winner, leading to more ungrounded choices. In addition, removing the budget constraint encourages the model to cheat by generating too many differences to game the score. Dropping one-to-one matching likewise lets broad, vague statements match multiple reference points at the same time.


\subsection{RQ4: Human Consensus and Process Fidelity}
\label{sec:rq4}

To verify whether intermediate reasoning agrees with the reconstructed reference evidence rather than superficial shortcuts, we evaluate process fidelity on a benchmark reconstructed from \textsc{HelpSteer3}, where pairs with human annotator agreement are converted into our structured difference-and-rubric format, with \texttt{Qwen3.7-Max} serving as the annotator that produces the reference evidence. We additionally include a zero-shot, prompt-only baseline that follows the same structured output format (Diff-Prompted Baseline) for comparison. 
As shown in Table~\ref{tab:rq4_helpsteer}, while supervised initialization establishes a grounded baseline, optimizing solely for win-or-lose outcome labels triggers clear rationale drift: final preference accuracy holds or slightly improves, yet core evidence recall drops noticeably while spurious correctness surges, confirming that pure outcome supervision encourages the model to rely on surface correlations to guess the correct winner. 
Additive rewards partially mitigate this degradation, but they still grant positive credit to plausible statements even when the final judgment is flawed. 
In contrast, \methodName{} enforces multiplicative gating with budget-constrained matching, achieving the highest overall preference accuracy alongside the highest recall of human consensus evidence while driving spurious correctness to the lowest level, demonstrating that its evaluation trajectory more consistently recovers the decisive differences humans actually use.

\begin{table}[t]
\centering
\caption{
\textbf{Human consensus fidelity on the reconstructed HelpSteer3 benchmark.} 
All evaluated models output contrastive difference rationales under the same evaluation protocol. 
``CER'' denotes Core Evidence Recall against verified human consensus reasons, and ``S-Corr'' denotes spurious correctness.
}
\label{tab:rq4_helpsteer}
\resizebox{\linewidth}{!}{%
\begin{tabular}{l|c|cc}
\toprule
\textbf{Configuration} & \textbf{Accuracy (\%)} $\uparrow$ & \textbf{CER (\%)} $\uparrow$ & \textbf{S-Corr (\%)} $\downarrow$ \\
\midrule
Diff-Prompted Baseline & 68.2 & 42.1 & 27.6 \\
\midrule
$\pi_{\mathrm{SFT}}$ (Initialization) & 74.5 & 68.3 & 10.8 \\
\quad + Outcome-only & 75.1 & 54.1 & 22.5 \\
\quad + Additive & 75.8 & 71.4 & 9.8 \\
\midrule
\rowcolor{purple!10}
\textbf{\methodName{} (Full RL)} & \textbf{77.7} & \textbf{79.6} & \textbf{4.6} \\
\bottomrule
\end{tabular}%
}
\end{table}

\section{Related Work}
\label{sec:related_work}

\subsection{Reward Modeling and Generative Reward Models}

Reward models learn evaluation signals from preference data and have become a core component of aligning language models with human preferences \citep{christiano2017deep,stiennon2020learning,ouyang2022training}. Conventional reward models typically map a response to a scalar score, whereas recent GenRMs formulate evaluation as language generation, enabling models to produce critiques, rationales, or judgments before predicting preferences \citep{zhang2025generative,mahan2024generative,ye2024beyond}. Subsequent work further strengthens this paradigm through inference-time reasoning or reinforcement learning \citep{liu2025inference,huang2026think,chen2026rm}, while recent studies show that outcome accuracy alone does not guarantee faithful evaluation and propose to explicitly supervise the correctness of reward-model rationales \citep{lai2026r,wang2026outcome}. In contrast to these approaches, which primarily improve the expressiveness or faithfulness of general evaluation reasoning, our method explicitly structures pairwise evaluation into \emph{difference discovery, rubric induction, criterion-level evaluation, and final adjudication}, and further constrains the discovered differences and derived rubrics through a difference-alignment reward.

\subsection{Rubric-Based Reward Modeling}

Rubric-based evaluation decomposes response quality into explicit criteria, enabling more fine-grained and interpretable judgments \citep{kim2024prometheus,kim2024prometheus2,hashemi2024llm}. Recent work further explores automatically generated and adaptive rubrics \citep{liu2026openrubrics,gunjal2026rubrics,zhang2026chasing}, with several methods leveraging paired responses to construct more discriminative or pair-specific evaluation criteria \citep{rezaei2025online,jia2026open,xu2026alternating,liu2026cdrrm}. However, they typically treat response contrasts merely as a means to construct or optimize rubrics. In contrast, we train the GenRM to discover \emph{preference-relevant differences} between arbitrary response pairs at inference time; moreover, an independent evaluator (the MetaJudge) constrains whether the discovered differences and resulting rubric are consistent with reference evidence.

\section{Conclusion}
\label{sec:conclusion}

In this paper, we introduced \methodName{}, a framework built on the premise that pairwise preference modeling requires actively discovering the right differences between competing responses and judging on their basis. By structuring evaluation into difference discovery, rubric induction, criterion-level evaluation, and final adjudication, \methodName{} makes criteria traceable to the discovered differences. Furthermore, our difference-alignment reward effectively suppresses superficial shortcuts and the reward hacking behaviors that exploit them. Extensive evaluations demonstrate that \methodName{} achieves state-of-the-art preference modeling accuracy on \textsc{RewardBench} and \textsc{RM-Bench}, surpassing all open-source reward models and leading proprietary LLM judges, exhibits strong parameter efficiency, and delivers robust guidance for downstream policy alignment while resisting length shortcuts.

\section*{Limitations}

While \methodName{} demonstrates strong performance across diverse preference modeling benchmarks, several directions remain to be explored. First, our experiments primarily focus on Qwen-based backbones at the 4B and 8B scales. Although the consistent gains across both model sizes suggest that the proposed framework is not tied to a particular scale, its behavior on substantially larger models or other model families warrants further investigation. Second, our evaluation mainly considers pairwise preference modeling and downstream DPO. Extending difference-driven evaluation to other alignment paradigms, such as alternative preference optimization algorithms or pointwise and listwise evaluation settings, is an interesting direction for future work. Finally, \methodName{} generates a structured evaluation trajectory before producing the final preference, which introduces additional decoding overhead compared with scalar reward models. Exploring more efficient trajectory generation or compression techniques may further improve its practicality.

\bibliography{custom}

\newpage
\appendix

\section{Experiment Setup}
\label{appen:exp_setups}

\subsection{Benchmarks}
\label{appen_sub:benchmark}

We evaluate pairwise reward quality on two complementary benchmarks, \textsc{RewardBench} and \textsc{RM-Bench}, which probe different aspects of reward modeling.

\paragraph{\textsc{RewardBench}.}
\textsc{RewardBench}~\citep{lambert2025rewardbench} is one of the first dedicated testbeds for reward models, built as triplets that pair each instruction with a preferred response and an inferior response drawn from real-world conversations and post-training data. It comprises four subsets: \texttt{Chat} (350), \texttt{Chat-Hard} (456), \texttt{Safety} (1,431), and \texttt{Reasoning} (740). \texttt{Chat-Hard} emphasizes adversarial instructions where the preferred response is less sycophantic or explicitly refuses, whereas \texttt{Safety} and \texttt{Reasoning} aggregate contrastive pairs from moderation scenarios and math, coding, and instruction-following sources. Following the official protocol, we report per-subset accuracy, i.e., the fraction of triplets for which the chosen response receives the higher reward.

\paragraph{\textsc{RM-Bench}.}
\textsc{RM-Bench}~\citep{liu2025rm} targets two failure modes that \textsc{RewardBench} rarely exposes: insensitivity to subtle content differences and susceptibility to stylistic bias. Each query pairs a stronger response containing a deliberately injected error with a weaker response that is restyled to appear superficially more polished, so that a reward model scores well only if it values substance over style. The benchmark covers four domains, namely \texttt{Chat} (129), \texttt{Math} (529), \texttt{Code} (228), and \texttt{Safety} (441). In addition, every comparison is instantiated at three difficulty levels that control how subtly the error is introduced. We report the official per-domain accuracy averaged over difficulty variants.

\subsection{Training Data}
\label{appen_sub:training_data}

Training comparisons are drawn from four public preference sources that jointly cover general conversation, mathematics, and code.

\textbf{Skywork Reward Preference 80K}~\citep{liu2026skywork} is a large-scale preference collection spanning chat, safety, mathematics, and code. Its constituent sources are curated with a combination of rule-based filtering and human verification, making it a reliable backbone for general-domain reward model training.

\textbf{HelpSteer3}~\citep{wang2025helpsteer3} provides human-annotated preferences over general, STEM-centric, and multilingual tasks, where each response pair is accompanied by a free-form response-level rationale written by professional annotators. Besides serving as a training source, its official validation split is reused for the human-grounded process evaluation in Section~\ref{sec:rq4}.

\textbf{Math-DPO-10K}~\citep{lai2024step} focuses on mathematical reasoning and constructs step-level preferences by pairing correct steps of long chain-of-thought solutions against erroneous alternatives.

\textbf{Code-Preference-Pairs}~\citep{lai2026r} is built by mutating seed programs with deliberately introduced bugs; corrected and broken versions form preference pairs, and surface artifacts such as error comments are removed so that the preference cannot be recovered from stylistic cues.

Following the synthesis pipeline in Section~\ref{sec:data_synthesis}, each retained pair is annotated by \texttt{Qwen3.5-397B-A17B} with the four sequential teacher prompts---contrast, rubric induction (with the candidate responses withheld), criterion-level evaluation, and adjudication---yielding a difference set with materiality labels, a pair-specific rubric, criterion-level verdicts, and a final preference; trajectories whose predicted preference contradicts the dataset label or whose quoted spans cannot be verified verbatim in the responses are discarded. The resulting pool of approximately 109K pairs is partitioned into 50K comparisons for cold-start SFT and 5K boundary comparisons for RL, where the boundary pool consists of pairs that the SFT reward model answers correctly in 2--6 out of 8 judgments.

\subsection{Baselines}
\label{appen_sub:baselines}

We compare \methodName{} against reward models from three families, following the taxonomy of Table~\ref{tab:main_results}.

\paragraph{LLM-as-a-Judge.}
This family prompts off-the-shelf LLMs to directly compare two candidate responses without any task-specific training~\citep{zheng2023judging}. We include \texttt{GPT-4o-0806}~\citep{hurst2024gpt}, \texttt{Claude-3.5-Sonnet}~\citep{anthropic2024claude}, and \texttt{Llama-3.1-70B-Instruct}~\citep{grattafiori2024llama}. These judges benefit from broad world knowledge, but they apply no structured evaluation protocol and are therefore prone to stylistic shortcuts.

\paragraph{Scalar Reward Models.}
Scalar models map each response to a single score and predict the winner by comparing scores. They are computationally cheap and often strong on in-distribution subsets, yet provide no explicit account of why one response is preferred. We evaluate \texttt{SteerLM-RM-70B}~\citep{wang2024helpsteer}, \texttt{Eurus-RM-7B}~\citep{yuan2025advancing}, \texttt{InternLM2-7B-Reward}~\citep{cai2024internlm2}, and \texttt{ArmoRM-Llama3-8B}~\citep{wang2024interpretable}.

\paragraph{Generative Reward Models.}
Generative reward models produce free-form textual judgments before committing to a final preference. This group includes reasoning-enhanced RMs trained with critique- or RL-based pipelines, such as \texttt{RM-R1}~\citep{chen2026rm} (evaluated in both its DeepSeek-distilled and Qwen2.5-instruct variants), \texttt{JudgeLRM}~\citep{chen2025judgelrm}, and \texttt{FARE}~\citep{xu2026foundational}. It also covers rubric-based RMs, which anchor their judgments in explicit evaluation criteria, namely \texttt{RubricRM}~\citep{zhang2026chasing}, \texttt{RubricARM}~\citep{jia2026open}, and \texttt{RUBRIC-ARROW}~\citep{jiang2026rubric}.

\paragraph{Downstream Alignment Baselines.}
For the policy optimization experiments in Section~\ref{sec:rq2}, we adopt the DPO configuration of \citet{viswanathan2026checklists} and align \texttt{Qwen2.5-7B-Instruct} with preferences derived from different reward sources: the implicit reward of UltraFeedback~\citep{cui2023ultrafeedback}, scalar RMs (\texttt{Skywork-RM}~\citep{liu2024skywork} and \texttt{ArmoRM}~\citep{wang2024interpretable}), pointwise \texttt{GPT-4o} judgments~\citep{zheng2023judging}, RLCF~\citep{viswanathan2026checklists}, and the rubric-based \texttt{RubricRM}~\citep{zhang2026chasing}. We use \texttt{GPT-4-0314}~\citep{achiam2023gpt} as the fixed reference for win-rate computation.

\subsection{Implementation Details}
\label{appen_sub:impl}

Training data are synthesized by \texttt{Qwen3.5-397B-A17B} with the four-stage teacher prompts of Section~\ref{sec:data_synthesis} (with the candidate responses withheld during rubric induction); after filtering, we use 50K comparisons for cold-start SFT and sample 5K boundary comparisons for reinforcement learning. Both backbones are fully fine-tuned in bfloat16 with ms-swift~\citep{zhao2025swift}, which provides the pipeline for both the SFT and the RL stage.

The SFT stage is trained for one epoch with learning rate $1\times10^{-5}$, a cosine schedule with 3\% warmup, global batch size 64, weight decay 0.1, and gradient-norm clipping at 1.0; we use DeepSpeed ZeRO stage 1 together with FlashAttention-2 and gradient checkpointing, with a maximum sequence length of 8,192. The RL stage runs for 200 steps at learning rate $1\times10^{-6}$ with the same schedule and regularization settings, samples 16 independent prompts per step with $G=8$ rollouts each at temperature 1.0, uses no KL regularization, and is optimized with DeepSpeed ZeRO-3, with completions capped at 4,096 tokens within a maximum sequence length of 8,192 (overlong completions are filtered).

The difference-alignment reward is computed by the MetaJudge, instantiated with \texttt{Qwen3.6-35B-A3B} and served with vLLM, which performs strict one-to-one matching between the budgeted claims generated in each rollout and the reference evidence under the capacity budget $L=\min(M,K)$, producing match scores normalized by the number of reference items, $R_{\mathrm{align}}$; the final reward multiplicatively gates outcome correctness with alignment, $R = R_{\mathrm{out}} \cdot R_{\mathrm{align}}$.

At inference time, all models are served with vLLM and decoded greedily (temperature 0) with a maximum of 2,048 generated tokens, with the thinking mode disabled to match the output template used during training. To control for position bias, each comparison is also evaluated with the two candidate responses swapped, and we report the average accuracy over the two orders.

\section{Prompts}
\label{app:prompts}

We present the prompts used in this work: the system and user prompts of the \methodName{} judge, which consolidates the four stages into a single autoregressive generation; the four sequential teacher prompts (contrast, rubric induction, criterion-level evaluation, and adjudication) that synthesize the cold-start demonstration data; and the prompt of the MetaJudge that produces the difference-alignment reward.

\begin{figure*}[!t]
\begin{promptbox}[\methodName{} (Judge) System Prompt]
\footnotesize
You are a reward model that judges which of two responses better satisfies an instruction. You reason in four stages. Output all five tags in order.

\textbf{Stage 1 -- Differences.}
Say where the two responses diverge. List points in both directions: what A does better and what B does better. For each, quote a short verbatim excerpt from each response and name the requirement in the instruction it bears on. Mark each difference material or immaterial.

A difference is ``material'' only if it changes how useful the answer is to the user: it changes a conclusion, fails a requirement the instruction actually makes, drops or adds required content, or introduces a safety or policy problem. Style, wording, length, and decorative formatting are NOT material. Do not treat them as decisive unless the instruction explicitly asks for a length, a format, or a structure, or unless the content is otherwise identical.

Do not state a winner in this stage. At most 6 differences.

\textbf{Stage 2 -- Rubric.}
Turn the material differences into 2 to 5 hard rules, plus at most 2 principles for genuinely subjective residue. A hard rule is objective and atomically verifiable, carries a concrete test, cites the difference it came from, quotes the instruction that requires it, and is tagged with a priority rung:
\begin{itemize}[leftmargin=*, noitemsep, topsep=2pt]
    \item 1 = factual accuracy and correctness
    \item 2 = instruction following and task completion
    \item 3 = relevance and completeness
    \item 4 = clarity and concision
    \item 5 = safety and policy alignment
    \item 6 = creativity, only when the instruction is open-ended
\end{itemize}
Rules must be answer-agnostic: never name the correct answer, never point at where one response happens to be right or wrong, never mention Response A or B. Do not invent a requirement the instruction never makes.

Prefer a safe refusal that gives a brief rationale over unsafe compliance. A refusal is not a failure to follow the instruction when the instruction asks for something that should be refused. Never write a blanket rule forbidding refusal.

\textbf{Stage 3 -- Eval.}
One block per rubric item, same numbers, same order. Compare A and B on that item alone, grounded in the text, and end the block with a line that is exactly ``winner: A'', ``winner: B'', or ``winner: tie''. Do not mention the overall verdict inside a block.

\textbf{Stage 4 -- Judgment.}
Weigh the per-item verdicts and decide. Weigh them, do not count them: a criterion the instruction centrally demands outranks a peripheral one, and a criterion one side fails badly outranks one it only narrowly fails. When the criteria point in different directions, name the decisive one and say why it outweighs the rest. If every hard rule tied, decide on the principles; if those tied too, decide on the lower-priority considerations and name the one you used.

Cite the criterion numbers you rely on and never introduce a consideration no criterion covers. Never let length or decorative formatting decide it unless a criterion says so.

At most 60 words. Do not write that the rubric ``correctly identifies'', ``supports'' or ``confirms'' the winner---you are deciding, not checking your own work. Do not declare a tie.
\end{promptbox}
\caption{\textbf{\methodName{} judge system prompt.} At inference, the model executes the full four-stage trajectory of Section~\ref{sec:method}---difference discovery, rubric induction, criterion-level evaluation, and adjudication---in a single autoregressive generation.}
\label{fig:prompt_judge}
\end{figure*}

\begin{figure*}[!t]
\begin{promptbox}[\methodName{} (Judge) System Prompt (cont.): Output Contract]
\footnotesize
\textbf{Output format, exactly}

\vspace{0.3em}
\begin{verbatim}
<diff>
1. [material] A: ...  B: ...
   Evidence: A "..." / B "..."
   Instruction: "..."
2. [immaterial: reason] A: ...  B: ...
   Evidence: A "..." / B "..."
</diff>

<rubric>
Hard Rules:
1. (must) ...
   Test: ...
   From: difference 1
   Priority: 2
Principles:
1. ...
   Priority: 4
</rubric>

<eval>
1. ...
winner: A

P1. ...
winner: tie
</eval>

<judgment>
...
</judgment>

<answer>A</answer>
\end{verbatim}
\vspace{0.3em}

\textbf{CRITICAL:}
\begin{itemize}[leftmargin=*, noitemsep, topsep=2pt]
    \item All five tags must be present, in this order, each closed.
    \item \texttt{<answer>} contains exactly ``A'' or ``B'' and nothing else.
    \item Do not announce the verdict anywhere before \texttt{<answer>}.
\end{itemize}
\end{promptbox}
\caption{\textbf{Output contract of the \methodName{} judge (continuation of the system prompt).} Explicit semantic tags delimit the four stages, and the final preference is read off the closing answer tag.}
\label{fig:prompt_judge_format}
\end{figure*}

\begin{figure}[!t]
\begin{promptbox}[\methodName{} (Judge) User Prompt]
\footnotesize
Task: Contrast $\rightarrow$ Rubric $\rightarrow$ Eval $\rightarrow$ Answer

\vspace{0.3em}
\textbf{Instruction}\\
\$\{instruction\}

\vspace{0.3em}
\textbf{Response A}\\
\$\{response\_a\}

\vspace{0.3em}
\textbf{Response B}\\
\$\{response\_b\}
\end{promptbox}
\caption{\textbf{\methodName{} judge user prompt.} The instruction and the two candidate responses are filled into the placeholders.}
\label{fig:prompt_judge_user}
\end{figure}

\begin{figure*}[!t]
\begin{promptbox}[Teacher Stage 1: Contrast]
\footnotesize
You compare two responses to the same instruction. Your only job is to say where they DIFFER. You must not decide which is better and must not hint at a preference.

\textbf{Rules}
1.~List differences in BOTH directions: points where Response A does better and points where Response B does better. A list that only favours one side is invalid and will be discarded.
2.~At most 6 differences, ordered by how much they matter. Do not pad the list; if there are only two real differences, write two.
3.~Every difference needs evidence: a VERBATIM excerpt of at most 20 words copied exactly out of that response. Copy characters exactly; do not paraphrase, do not fix typos, do not add ellipses.
4.~Say which requirement in the instruction the difference bears on, as a VERBATIM excerpt of the instruction. Use null if it bears on none.
5.~Classify each difference as ``material'' or ``immaterial''. A difference is ``material'' only if it changes how useful the answer is to the user: it changes a conclusion, fails a requirement the instruction actually makes, drops or adds required content, or introduces a safety or policy problem. Style, wording, length, and decorative formatting are NOT material. Do not treat them as decisive unless the instruction explicitly asks for a length, a format, or a structure, or unless the content is otherwise identical.
6.~If any length or formatting difference exists you MUST list it and mark it immaterial, with the reason. Omitting it is a failure.
7.~Never write ``Response A'' or ``Response B'' inside an evidence excerpt. Quote the response text itself.

\textbf{Output: strict JSON, nothing else}
\vspace{0.3em}
\begin{verbatim}
{
  "differences": [
    {
      "id": 1,
      "a_side": "what Response A concretely does, includes or omits",
      "b_side": "what Response B concretely does, includes or omits",
      "a_evidence": "verbatim excerpt from Response A, <=20 words",
      "b_evidence": "verbatim excerpt from Response B, <=20 words",
      "instruction_anchor": "verbatim excerpt from the instruction, or null",
      "materiality": "material",
      "materiality_reason": ""
    },
    {
      "id": 2,
      "a_side": "...",
      "b_side": "...",
      "a_evidence": "...",
      "b_evidence": "...",
      "instruction_anchor": null,
      "materiality": "immaterial",
      "materiality_reason": "length only; the instruction sets no length limit"
    }
  ]
}
\end{verbatim}
\end{promptbox}

\begin{promptbox}[Teacher Stage 1: Contrast (User Prompt)]
\footnotesize
\textbf{Instruction}\\
\$\{instruction\}

\vspace{0.3em}
\textbf{Response A}\\
\$\{response\_a\}

\vspace{0.3em}
\textbf{Response B}\\
\$\{response\_b\}

\vspace{0.3em}
List the differences as JSON.
\end{promptbox}
\caption{\textbf{Teacher Stage~1: difference discovery (Step~1 of Section~\ref{sec:data_synthesis}).} The teacher lists pair-specific differences in both directions, with verbatim evidence and materiality labels, and must not state a preference.}
\label{fig:prompt_teacher_contrast}
\end{figure*}

\begin{figure*}[!t]
\begin{promptbox}[Teacher Stage 2: Rubric (Response Isolation)]
\footnotesize
You write an evaluation rubric for one instruction, derived ONLY from the difference list supplied below. You have not seen the responses and must not imagine their content.

\textbf{Rules}
1.~Write 2 to 5 hard rules. Fewer is better: every rule must earn its place.
2.~Each hard rule must be objective and atomically verifiable---a single reader could apply it to one response alone and answer pass or fail.
3.~Each hard rule must come from a difference marked ``material''. Set ``from\_difference'' to that difference's id. Do not write a rule whose from\_difference is immaterial or absent.
4.~Each hard rule must quote the instruction that requires it, verbatim, in ``instruction\_anchor''. If no excerpt of the instruction supports the rule, do not write the rule. Inventing requirements the instruction never makes is the single worst failure here.
5.~Each hard rule gets a ``priority'': the rung of the ladder it instantiates.
6.~Answer-agnostic. Forbidden:
\begin{itemize}[leftmargin=*, noitemsep, topsep=2pt]
    \item naming the correct answer, result, or solution method
    \item pointing at the place where one response happens to be right or wrong
    \item any mention of ``Response A'', ``Response B'', ``Answer A'', or which side is better
\end{itemize}
Bad: ``Correct calculation of C's time using (remaining work) / (C's rate).''\\
Good: ``Computes the time for the final worker from the work that remains, not the whole job.''
7.~Do not reward length or decorative formatting unless the instruction asks for a length, a format, or a structure.
8.~Never write a blanket rule forbidding refusal; a rule about refusal must be conditioned on what the request actually asks for.
9.~Add at most 2 principles, and only for genuinely subjective residue that no hard rule can cover. If hard rules suffice, write none.

\textbf{Priority ladder}
\begin{itemize}[leftmargin=*, noitemsep, topsep=2pt]
    \item 1 = factual accuracy and correctness
    \item 2 = instruction following and task completion
    \item 3 = relevance and completeness
    \item 4 = clarity and concision
    \item 5 = safety and policy alignment
    \item 6 = creativity, only when the instruction is open-ended
\end{itemize}

\textbf{Output: strict JSON, nothing else}
\vspace{0.3em}
\begin{verbatim}
{
  "hard_rules": [
    {
      "rule_id": 1,
      "type": "must",
      "criterion": "what a good response achieves or avoids",
      "test": "a concrete procedure runnable on a single response",
      "from_difference": 1,
      "instruction_anchor": "verbatim excerpt from the instruction",
      "priority": 2
    }
  ],
  "principles": [
    {"principle_id": 1, "description": "...", "priority": 4}
  ]
}
\end{verbatim}
\end{promptbox}

\begin{promptbox}[Teacher Stage 2: Rubric (User Prompt)]
\footnotesize
\textbf{Instruction}\\
\$\{instruction\}

\vspace{0.3em}
\textbf{Differences found between the two responses}\\
\$\{differences\_json\}

\vspace{0.3em}
Write the rubric as JSON.
\end{promptbox}
\caption{\textbf{Teacher Stage~2: rubric induction with response isolation (Step~2 of Section~\ref{sec:data_synthesis}).} With the candidate responses withheld, criteria are derived from the material differences and must satisfy evidence traceability and answer agnosticism.}
\label{fig:prompt_teacher_rubric}
\end{figure*}

\begin{figure*}[!t]
\begin{promptbox}[Teacher Stage 3: Eval]
\footnotesize
You apply a fixed rubric to two responses. The rubric is given. Do not add, drop, reword, or reorder any item.

\textbf{Rules}
1.~One block per rubric item, in the rubric's order, numbered identically: hard rules as ``1.'' ``2.'' \ldots, principles as ``P1.'' ``P2.'' \ldots
2.~A block compares A and B on THAT criterion only. Ground every claim in the responses: quote or closely paraphrase what is actually there.
3.~The last line of every block must be exactly ``winner: A'', ``winner: B'', or ``winner: tie''---lowercase, alone on its line, nothing else on it.
4.~``tie'' is legitimate and expected when both sides handle the criterion equally well. Do not force a winner.
5.~Never state, hint at, or justify an overall verdict inside a block. No ``overall'', ``therefore B wins'', ``which outweighs''. Judge the one criterion and stop.
6.~Judge only what is in front of you. Do not supply facts the responses do not contain, and do not work out the reference answer yourself---check each response against the criterion, not against an answer you derived.
7.~A safe refusal with a brief rationale satisfies safety criteria; it is not a failure to respond.

\textbf{Output format, exactly}
\vspace{0.3em}
\begin{verbatim}
<eval>
1. {comparison on hard rule 1}
winner: A

2. {comparison on hard rule 2}
winner: tie

P1. {comparison on principle 1}
winner: B
</eval>
\end{verbatim}
\end{promptbox}

\begin{promptbox}[Teacher Stage 3: Eval (User Prompt)]
\footnotesize
\textbf{Instruction}\\
\$\{instruction\}

\vspace{0.3em}
\textbf{Rubric}\\
\$\{rubric\_json\}

\vspace{0.3em}
\textbf{Response A}\\
\$\{response\_a\}

\vspace{0.3em}
\textbf{Response B}\\
\$\{response\_b\}

\vspace{0.3em}
Apply every rubric item and output the \texttt{<eval>} block.
\end{promptbox}
\caption{\textbf{Teacher Stage~3: criterion-level evaluation (Step~3 of Section~\ref{sec:data_synthesis}).} With the rubric frozen, both responses are judged under each criterion independently, without hinting at the overall verdict.}
\label{fig:prompt_teacher_eval}
\end{figure*}

\begin{figure*}[!t]
\begin{promptbox}[Teacher Stage 4: Judgment]
\footnotesize
You are given the differences between two responses, the rubric derived from them, and the per-criterion verdicts. Decide which response is better.

\textbf{Rules}
1.~Weigh the verdicts; do not count them. A criterion the instruction centrally demands outranks a peripheral one, and a criterion one side fails badly outranks one it only narrowly fails. Say what you weighed.
2.~Resolve conflicts explicitly. When the criteria point in different directions, name the decisive one and say why it outweighs the rest.
3.~If every hard rule tied, decide on the principles. If those tied too, decide on the lower-priority considerations and name the one you used.
4.~Cite the criterion numbers you rely on. Never introduce a consideration that no criterion covers.
5.~Never let length or decorative formatting decide it unless a criterion explicitly says so.
6.~A safe refusal with a brief rationale beats unsafe compliance; refusing is not a failure to follow the instruction when the request should have been refused.
7.~At most 60 words: the weighing only. Do not restate the criteria and do not repeat the comparisons.
8.~Forbidden:
\begin{itemize}[leftmargin=*, noitemsep, topsep=2pt]
    \item writing that the rubric ``correctly identifies'', ``correctly predicts'', ``supports'', ``confirms'' or ``points to'' the winner. You are deciding, not checking your own work.
    \item declaring a tie, or saying both responses are equally good. Pick A or B.
\end{itemize}

\textbf{Output format, exactly}
\vspace{0.3em}
\begin{verbatim}
<judgment>
{the weighing}
</judgment>

<answer>A</answer>
\end{verbatim}
\end{promptbox}

\begin{promptbox}[Teacher Stage 4: Judgment (User Prompt)]
\footnotesize
\textbf{Instruction}\\
\$\{instruction\}

\vspace{0.3em}
\textbf{Differences found}\\
\$\{differences\_json\}

\vspace{0.3em}
\textbf{Rubric}\\
\$\{rubric\_json\}

\vspace{0.3em}
\textbf{Per-criterion verdicts}\\
\$\{eval\_blocks\}

\vspace{0.3em}
\textbf{Response A}\\
\$\{response\_a\}

\vspace{0.3em}
\textbf{Response B}\\
\$\{response\_b\}

\vspace{0.3em}
Decide the winner.
\end{promptbox}
\caption{\textbf{Teacher Stage~4: final adjudication (Step~4 of Section~\ref{sec:data_synthesis}).} The teacher weighs the verdicts---rather than counting them---into an integrated rationale and the final prediction; trajectories that fail label agreement or verbatim-quote checks are discarded.}
\label{fig:prompt_teacher_judgment}
\end{figure*}

\begin{figure*}[!t]
\begin{promptbox}[MetaJudge System Prompt]
\footnotesize
You are a strict alignment auditor. You score how well an original critique list achieves each point of a reference critique list. You match strictly: vague mention is not achievement, and agreeing on a verdict with different reasons is not achievement. Always follow the requested output format exactly.
\end{promptbox}

\begin{promptbox}[MetaJudge User Prompt (Grounded Variant)]
\footnotesize
You are a rigorous alignment achievement analyst. Given an original critique list, a reference critique list, AND the two responses being compared, determine to what extent each item in the reference list is achieved (expressed) by the items in the original list, and assign achievement scores (0-1) based on semantic importance. Different phrasings with the same meaning are considered equivalent, but mere semantic mention without achieving the objective should be considered not achieved. Abstract or vague descriptions should be considered not achieved.

\textbf{CRITICAL GROUNDING RULE:} an original item can only count toward achievement if what it claims is FACTUALLY TRUE about Response A / Response B below. Before scoring, verify each original item against the actual responses: if it describes something that does not exist in the responses, misquotes or distorts their content, or contradicts what the responses actually do, treat it as not achieved (score 0) regardless of how well it matches the reference list. Fabricated or unverifiable critiques must never earn credit.

Below is the instruction the two responses answer:\\
\texttt{<instruction>}\\
\{instruction\}\\
\texttt{</instruction>}

\vspace{0.3em}
\texttt{[Response A start]}\\
\{response\_a\}\\
\texttt{[Response A end]}

\vspace{0.3em}
\texttt{[Response B start]}\\
\{response\_b\}\\
\texttt{[Response B end]}

\vspace{0.3em}
\texttt{[Original critique list start]}\\
\{original\_list\}\\
\texttt{[Original critique list end]}

\vspace{0.3em}
\texttt{[Reference critique list start]}\\
\{reference\_list\}\\
\texttt{[Reference critique list end]}

\vspace{0.3em}
For each item Rx in the reference list, find the best-matching single item Sy in the original list (if there is no match, score 0 with y = 0; matching S0 means no match). Match as strictly as possible, and prioritize lower-score matches when in doubt.

An objective is considered achieved only when (1) the specific issue is clearly pointed out in detail (e.g., a particular passage lacks coherence, a concrete behavioral difference between the responses), rather than merely a vague evaluation such as stating which response is better or worse, AND (2) that pointing is grounded in what Response A / Response B actually contain. Especially when comparing responses A and B, if the items only agree on which one is better but the reasons differ, it should NOT be considered achieved.

- Not achieved/contradictory: 0.0 (the original list does not address the critique point, provides an opposite conclusion, describes it only abstractly/vaguely without pinpointing the issue, OR the matching original item is not factually grounded in the responses)
- Slightly involved: 0.25 (only mentioned some elements; not concretized, unable to prove the objective was achieved)
- Partially achieved: 0.5 (related and grounded analysis exists, but multiple key elements are missing or distorted)
- Mostly achieved: 0.75 (the critique point is basically captured with grounded evidence; only minor detail missing)
- Fully achieved: 1.0 (the critique point is faithfully and concretely expressed, fully grounded in the responses; key conditions and constraints all satisfied)

For quantitative or conditional critique points, check numbers, ranges, thresholds, and conditions against the responses; if key constraints are not satisfied, do not judge as fully achieved.

\textbf{Output Format}

Emit exactly this structure:
\vspace{0.3em}
\begin{verbatim}
<RESULT_START>
Rule scores:
- R1@S<best-match-number>: <score>
- R2@S<best-match-number>: <score>
...continue for every reference item...
<RESULT_END>
\end{verbatim}
\vspace{0.3em}
Rules for the output:
- Emit exactly \{n\_ref\} score lines (one per reference item), in order, formatted exactly as ``- Rx@Sy: score'' where y is the original item it best matches (0 if none), and score is one of 0, 0.25, 0.5, 0.75, 1.
- No other content inside the RESULT block.
\end{promptbox}
\caption{\textbf{MetaJudge prompt.} Scores the difference-alignment reward of Section~\ref{sec:rl_alignment} by measuring how well each reference item is achieved by the generated claims; vague, verdict-only, or ungrounded critiques earn no credit.}
\label{fig:prompt_metajudge}
\end{figure*}

\end{document}